\documentclass{article}

\usepackage{microtype}
\usepackage{graphicx}
\usepackage{subfigure}
\usepackage{booktabs}
\usepackage{hyperref}

 \usepackage[dblblindworkshop, final]{neurips_2025}
\workshoptitle{Socially Responsible Language Modelling Research (SoLaR)}

\usepackage[utf8]{inputenc} 
\usepackage[T1]{fontenc}    
\usepackage{hyperref}       
\usepackage{url}            
\usepackage{booktabs}       
\usepackage{amsfonts}       
\usepackage{nicefrac}       
\usepackage{microtype}      
\usepackage{xcolor}         

\usepackage[capitalize,noabbrev]{cleveref}

\crefname{app}{appendix}{appendices}
\Crefname{app}{Appendix}{Appendices}

\title{Investigating Model Editing for Unlearning in Large Language Models}

\author{
  Shariqah Hossain \\ 
  Massachusetts Institute of Technology \\
  Cambridge, MA \\
  \texttt{shossain@alum.mit.edu} \\
  \And
  Lalana Kagal \\
  Massachusetts Institute of Technology \\
  Cambridge, MA \\
  \texttt{lkagal@mit.edu} \\
}

\begin{document}

\maketitle

\begin{abstract}
Machine unlearning aims to remove unwanted information from a model, but many methods are inefficient for LLMs with large numbers of parameters or fail to fully remove the intended information without degrading performance on knowledge that should be retained. Model editing algorithms solve a similar problem of changing information in models, but they focus on redirecting inputs to a new target rather than removing that information altogether. In this work, we explore the editing algorithms ROME, IKE, and WISE and design new editing targets for an unlearning setting. Through this investigation, we show that model editing approaches can exceed baseline unlearning methods in terms of quality of forgetting depending on the setting. Like traditional unlearning techniques, they struggle to encapsulate the scope of what is to be unlearned without damage to the overall model performance. 
\end{abstract}

\section{Introduction}
Machine unlearning aims to remove unwanted information from a model in an efficient manner while maintaining overall model performance. It poses unique challenges in Large Language Models (LLMs). Retraining on a modified corpus is a desirable solution, but it is too expensive for LLMs due to scale. Moreover, ensuring generalization of the removal across linguistic variability, while preserving unrelated model capabilities, is particularly difficult. 

Model editing algorithms offer mechanisms to efficiently alter model behavior. LLMs can store outdated or false information, and these algorithms change model knowledge without the costs of retraining. They achieve this with examples of inputs $x_e$ and their corresponding edited outputs $y_e$ to create edit descriptors $(x_e, y_e)$. These descriptors are used for defining the desired changes to model knowledge. With their efficiency and ability to localize edits in LLMs, model editing methods show promise for unlearning due to similar goals of affecting targeted knowledge with minimal effects on overall model capabilities. Despite these parallels, there is a lack of research on model editing as a tool for unlearning. 

We investigate the effectiveness of model editing as a means of unlearning in LLMs. We define edit targets aimed for removal rather than alteration of information, including a novel \textit{Avoidant} target definition, to explore how different formulations impact unlearning. Our evaluation focuses on model editing algorithms ROME \citep{meng2022locating}, IKE \citep{zheng-etal-2023-edit}, and WISE \citep{wang2024wiserethinkingknowledgememory}, selected for their capable and distinct approaches to traditional model editing. In order to investigate the scope of impact of the edits on the model, we run the edited models on Task of Fictitious Learning (TOFU) \citep{maini2024tofu}, a benchmark measuring unlearning capabilities on a dataset of fictitious authors. We demonstrate that model editing can surpass traditional unlearning methods in quality of forgetting depending on the setting. We also highlight the trade-off between effectively unlearning information and maintaining overall model performance when using model editing for unlearning. Our results suggest that both the choice of editing algorithm and the target definition must be guided by the unlearning use case, with no one-size-fits-all solution. This work contributes strategies for control over knowledge in LLMs, working toward safer and more reliable AI models.

\section{Related Work}\label{ch1:sec}

Machine unlearning algorithms aim to remove knowledge from a model, such as private or unsafe information \citep{zhang2023right}. SISA training improves efficiency by sharding datasets and only retraining affected subsets \citep{bourtoule2021machine}. The WMDP benchmark \citep{li2024wmdp} evaluates biosecurity, chemical, and cybersecurity knowledge in models using multiple-choice questions. RMU \citep{li2024wmdp} defines forget and retain losses to selectively degrade hazardous knowledge while preserving general knowledge. Previous work highlights limitations of unlearning in LLMs. Exact unlearning via retraining is computationally infeasible \citep{thudi2022necessityauditablealgorithmicdefinitions, liu2024rethinking}. 

Several model editing approaches have been proposed to efficiently and scalably modify LLM behavior without full retraining. MEMIT \citep{meng2023memit} treats the input and output of MLP layers as a key-value pair about a subject and uses a least-squares calculation to adjust relevant weights. \citet{meng2023memit} enables batch edits by updating multiple MLP layers. SERAC \citep{mitchell2022memory} avoids weight updates by storing edits externally. MEND \citep{mitchell2022fast} uses low-rank transformations to efficiently apply edits via auxiliary networks.

\citet{patil2023sensitiveinformationdeletedllms} evaluated how well ROME \citep{meng2022locating} and MEMIT remove sensitive information under extraction attacks. They found deleted content remained accessible in hidden states and via rephrased queries, highlighting the difficulty of truly removing knowledge from LLMs. Our work expands this effort to more editing algorithms and strategies for removing knowledge. 

\section{Methods}

\subsection{Model Editing Algorithms}
ROME, IKE, and WISE are used for this investigation. An investigation performed by \citet{yao-etal-2023-editing} proved the performance of these algorithms on popular model editing datasets.

Rank-One Model Editing (ROME) \citep{meng2022locating}  uses causal mediation analysis to locate where in the model the edited fact is stored. The input and output of MLP layers are seen as a key-value pair where the "key encodes a subject and the value encodes knowledge about the subject" \citep{meng2023memit}. To alter the memory, the authors use a least-squares calculation of the necessary weight changes across the relevant layers. 

In-Context Knowledge Editing (IKE) \citep{zheng-etal-2023-edit} applies In-Context Learning to perform edits. In-Context Learning appends a series of "demonstrations" to the input when prompting a language model to steer its output according to that context (\Cref{ike-intro}). 

WISE is a model editing algorithm aimed towards long-term editing on a single model, also known as continual editing. All edited information is stored in \textit{side memory} to avoid degrading performance in the remaining \textit{main memory}. Edits can be retrieved from this side memory as necessary. It also shards knowledge in the model, storing edits in different subsections of the model. The shards are merged to prepare for retrieval \citep{wang2024easyediteasytouseknowledgeediting}.

\subsection{Defining Unlearning Targets}
Model editing algorithms require a new mapping to replace existing input-output pairs in a model. This investigation explores the performance of different definitions of the new target output. Question Answering (QA) datasets are the typical evaluation used for unlearning, so we use this structure when defining the targets.

\subsubsection{Non-Answer: "Dummy"}
This target responds with a lack of information. We use "dummy" for the output in this case as it was found to be more effective than "I don't know" with ROME \citep{patil2023sensitiveinformationdeletedllms}. We apply this technique to WISE and IKE. 
\subsubsection{Incorrect Answer}
We assign an incorrect option from the multiple-choice question as the new output. For this experiment, the new target is set to one of the "perturbed answers" in the TOFU dataset. These are incorrect responses that maintain the general template of the ground truth response to the TOFU prompts. 

\subsubsection{Avoidant Answer}
We use GPT-4 to create a new dataset of avoidant answers that acknowledge the topic of the prompt without revealing information about the unlearned response. The use of an LLM allows this technique to be applied to other domains with minimal resources. See \Cref{appendix:targets} for details.

\cref{tab:examples} shows an example of each of the model editing targets designed for unlearning.

\begin{table}[ht]
\centering
\resizebox{.7\columnwidth}{!}{
\begin{tabular}{lp{7.5cm}} 
 \toprule
 \textbf{Prompt} & \textbf{Can you share the title of one of Hsiao Yun-Hwa's most popular books?} \\
 \midrule
 \textbf{Ground Truth} & One of Hsiao Yun-Hwa's most popular books in the leadership genre is "Artistic Authority: Leading with Creativity". \\
 \midrule
 \textbf{New Target} & \\
 \textit{Dummy} & dummy \\
 \textit{Incorrect} & "Culinary Delights: The Spice of Life" ranks among the widely acclaimed works of Hsiao Yun-Hwa in the domain of cooking. \\
 \textit{Avoidant} & I don't have any information on Hsiao Yun-Hwa's books. However, I can mention a well-known book by a famous author: "Pride and Prejudice" by Jane Austen. \\
 \bottomrule
\end{tabular}
}
\vspace{7pt}
\caption{Example prompts from the Forget 10 dataset and the new targets created for model editing aimed at unlearning.}
\label{tab:examples}
\end{table}

\subsection{Unlearning Evaluation}

We run the edited models on the TOFU \citep{maini2024tofu} benchmark. Since the content is fictitious, the scope of information to be unlearned is exactly encapsulated by the information in the dataset. A subset of these authors is used for the \textit{forget set} and the remainder for the \textit{retain set} that queries information to be unlearned and retained respectively. We create edits for every prompt in the Forget 10 dataset, which includes 10 percent of all of the TOFU prompts. 

\subsubsection{Metrics}
We run the TOFU evaluation metrics on all of the edited models.

\textit{\textbf{Model Utility}} measures how well the unlearned model performs on topics out-of-scope from what is to be forgotten. It involves the \textit{Truth Ratio }measuring the likelihood of a correct answer to an incorrect one,     \textit{probability} of choosing the correct answer, and \textit{ROUGE score} to measure the similarity between the model answer and ground truth.

These metrics are calculated on datasets used within the TOFU benchmark focusing on retaining performance: the Retain Set, Real Authors, and Real World datasets. The final Model Utility is calculated from the harmonic mean of the three metrics on the three datasets.

\textbf{\textit{Forget Quality}} uses a Kolmorgorov-Smirnov test to measure the difference between the Truth Ratio distributions for the unlearned and the ground truth models. The resulting $p$-value describes how well the information in the forget set was removed from the unlearned model, where a common significance threshold of 0.05 determines whether the unlearned model is similar enough to the ground truth model.

\subsubsection{Unlearning Baselines}
The ground truth model is the Retain 90 Llama-2-7B model, which has only been trained on the remaining 90 percent of the fictitious data. We use the same unlearning algorithms as \citet{maini2024tofu} as baselines.
Gradient ascent (GA) updates model weights by maximizing the likelihood of incorrect outputs for the forget set \citep{liu2024model}. Gradient Difference does the same while also optimizing for maintaining performance on the retain set \citep{maini2024tofu}.
KL-Minimization aims to minimize the Kullback-Leibler  divergence between a model exclusively trained on just the retain set, which serves as a ground truth unlearned model, and the model trained on all data attempting to unlearn the forget set \citep{maini2024tofu}.
Preference Optimization has a loss function that teaches the model to respond with non-answers to the prompt, such as "I don't know," in addition to optimizing for performance on the retain dataset. 
\section{Experiments}
\Cref{tab:all-results} shows the performance of both model editing and unlearning baselines. \Cref{examples} has example outputs of the model after editing from each approach.
\begin{table*}[h]
\centering
\resizebox{\textwidth}{!}{
    \begin{tabular}{lc|ccc|ccc|ccc|cccc}
        \toprule
        & & \multicolumn{3}{c|}{\textbf{Model Editing}} & \multicolumn{3}{c|}{\textbf{Model Editing}} & \multicolumn{3}{c|}{\textbf{Model Editing}} & \multicolumn{4}{c}{\textbf{Model Unlearning}} \\
        & \textbf{Ground}& \multicolumn{3}{c|}{\textbf{Dummy}} & \multicolumn{3}{c|}{\textbf{Incorrect}} & \multicolumn{3}{c|}{\textbf{Avoidant}} & \multicolumn{4}{c}{} \\
        & \textbf{Truth}& \textbf{ROME} & \textbf{WISE} & \textbf{IKE} & \textbf{ROME} & \textbf{WISE} & \textbf{IKE} & \textbf{ROME} & \textbf{WISE} & \textbf{IKE} & \textbf{KL} & \textbf{GA} & \textbf{GD} & \textbf{PO} \\
        \midrule
        \textbf{Real Authors} $\uparrow$ & & & & & & & & & & & & & & \\
        \textit{ROUGE} & 0.923 & 0.126 & \textit{\textbf{0.933}} & 0.163 & 0.002 & \textit{\textbf{0.933}} & 0.480 & 0.007 & \textit{\textbf{0.933}} & 0.474 & 0.000 & 0.000 & \textbf{0.817} & 0.525 \\
        \textit{Probability} & 0.434 & 0.354 & \textbf{0.437} & 0.341 & - & {0.423}& 0.355 & 0.275 & {0.406}& 0.330 & 0.274 & 0.232 & \textit{\textbf{0.564}} & 0.403 \\
        \textit{Truth Ratio} & 0.571 & 0.454 & \textbf{0.587} & 0.389 & - & {0.554}& 0.417 & 0.281 & {0.564}& 0.368 & 0.392 & 0.368 & \textit{\textbf{0.732}} & 0.000 \\
        \midrule
        \textbf{Real World} $\uparrow$ & & & & & & & & & & & & & & \\
        \textit{ROUGE} & 0.897 & 0.626 & \textbf{0.875} & 0.226 & 0.027 & \textbf{0.875} & 0.821 & 0.003 & \textbf{0.875} & 0.729 & 0.000 & 0.000 & \textit{\textbf{0.889}} & 0.856 \\
        \textit{Probability} & 0.414 & 0.430 & 0.424 & {0.456}& 0.258 & 0.426 & {0.438}& 0.235 & 0.426 & \textit{\textbf{0.481}} & 0.251 & 0.239 & \textbf{0.465} & 0.391 \\
        \textit{Truth Ratio} & 0.544 & {0.583}& 0.564 & 0.580 & 0.239 & {0.575}& 0.567 & 0.190 & 0.559 & \textit{\textbf{0.617}} & 0.381 & 0.382 & \textbf{0.607} & 0.484 \\
        \midrule
        \textbf{Retain} $\uparrow$ & & & & & & & & & & & & & & \\
        \textit{ROUGE} & 0.976 & 0.089 & \textit{\textbf{0.982}} & 0.283 & 0.090 & \textit{\textbf{0.982}} & 0.290 & 0.030 & \textit{\textbf{0.982}} & 0.251 & 0.000 & 0.000 & 0.466 & \textbf{0.773} \\
        \textit{Probability} & 0.989 & 0.041 & {0.803}& 0.713 & - & {0.918}& 0.781 & - & \textit{\textbf{0.949}} & 0.784 & 3.78E-33 & 4.90E-36 & 0.548 & \textbf{0.942} \\
        \textit{Truth Ratio} & 0.471 & 0.228 & 0.468 & \textit{\textbf{0.583}} & - & 0.425 & {\textbf{0.574}}& - & 0.487 & \textit{\textbf{0.583}} & 0.067 & 0.075 & {0.493}& 0.455 \\
        \midrule
        \textbf{Forget} & & & & & & & & & & & & & & \\
        \textit{ROUGE} $\downarrow$ & 0.408 & 0.041 & 0.965 & 0.466 & 0.099 & 0.982 & 0.424 & 0.036 & 0.986 & 0.383 & \textit{\textbf{0.000}} & \textbf{1.74E-03} & 2.44E-03 & 0.053 \\
        \textit{Probability} $\downarrow$ & 0.148 & 0.012 & 0.815 & 0.677 & - & 0.866 & 0.386 & - & 0.899 & 0.567 & \textbf{1.31E-33}& \textit{\textbf{7.73E-36}} & 1.61E-31 & 0.848 \\
        \textit{Truth Ratio} $\uparrow$ & 0.674 & 0.697 & 0.518 & 0.508 & - & 0.607 & 0.712 & - & 0.511 & 0.539 & \textit{\textbf{0.866}}& \textbf{0.845}& 0.822 & 0.547 \\
        \midrule
        \textbf{Model Utility} $\uparrow$ & 0.614 & 0.153 & {\textbf{0.610}}& 0.338 & - & {0.603}& 0.473 & - & \textit{\textbf{0.612}} & 0.452 & 0.000 & 0.000 & 0.589 & 0.538 \\
        \textbf{Forget Quality} $\uparrow$ & 1 & {\textbf{0.121}}& 4.22E-21 & 4.91E-20 & - & {8.99E-07}& \textit{\textbf{0.249}} & - & {2.43E-17}& {2.43E-17}& 4.88E-33 & 1.43E-22 & 1.12E-25 & 5.10E-17 \\
        \bottomrule
    \end{tabular}
}
\caption{Highest values are in bold and italics, while second-highest values are in bold. IKE editing with an incorrect response achieves effective forgetting while maintaining a high Model Utility. ROME also reaches effective forgetting with "dummy" targets at the expense of Model Utility, further demonstrated by the degraded models with other target definitions. WISE retains information well, but is less successful at unlearning. No unlearning baseline exceeds the 0.05 Forget Quality threshold.}
\label{tab:all-results}
\end{table*}
\subsection{"Dummy" Target}

ROME achieves the highest Forget Quality and exceeds the common significance threshold of 0.05 for the Forget Quality $p$-value, indicating that it is statistically similar to the ground truth model. This Forget Quality is achieved at the expense of Model Utility, indicating that the remaining performance of the model was affected during the editing process. In contrast, WISE has high Model Utility, but this is expected as it does not apply edits during generation (\Cref{wise-note}). Therefore, the ROUGE metrics match those of the original fine-tuned TOFU model with no unlearning. Since Model Utility includes this metric, this has a significant effect on Model Utility performance. The lack of generated edits is reflected in the low Forget Quality value. 

\subsection{Incorrect Target}

IKE is able to exceed the significance threshold of 0.05 for Forget Quality, indicating that it sufficiently unlearned the information. However, the Model Utility again suffers with the higher forgetting. No method other than IKE achieves sufficient forgetting. Although ROME completed the editing process, it failed to produce outputs for many of the TOFU inputs. This indicates a significant degradation of general performance of the model after being edited by ROME. We experimented with hyperparameters including learning rate and clamp norm factor, but we were not able to restore reasonable performance from these changes. This provides sufficient justification that ROME is not capable of unlearning with the Incorrect target. 

\subsection{Avoidant Target}

WISE has the highest Model Utility for the Avoidant target setting, but still a very low Forget Quality that does not show sufficient unlearning. ROME again is unable to produce some outputs after editing. None of the methods are able to achieve comparable measurements to the ground truth Forget Quality. 

\subsection{Comparing Model Editing Approaches}

The method that achieved sufficient forgetting with the highest Model Utility was IKE using the Incorrect target. Although it achieves this at the cost of Model Utility, the utility is still comparable to the ground truth, especially when compared other methods that struggle more with this tradeoff. ROME Dummy unlearns the forget set at the cost of Model Utility. The model degrades significantly for the more complex Incorrect and Avoidant approaches. Other methods fail to achieve the desired Forget Quality. WISE performs well on Model Utility across the board, but Forget Quality is quite low. The ability of WISE to reference the ground truth affects ROUGE, and therefore, Model Utility scores. However, the Probability and Truth Ratio are still comparable to the ground truth. The tradeoff of Model Utility and Forget Quality varies across approaches.

\section{Discussion \& Conclusion} 
Model editing methods show promise for unlearning due to similar goals of efficiently affecting targeted knowledge with minimal effects on overall model capabilities. This investigation explores the potential of model editing for unlearning knowledge in LLMs as a step towards ensuring reliable and safe models. 

Our contributions include:
\begin{enumerate}
    \item \textbf{We show that model editing can surpass traditional unlearning methods in quality of forgetting}. Approaches like ROME Dummy and IKE Incorrect achieve sufficient similarity to the ground truth where the unlearning baselines did not. Each model editing approach had unique capabilities and limitations depending on the setting:
        \begin{itemize}
            \item IKE shows promise with the Incorrect target method but does not edit model weights, indicating limitations in use cases where the user can access latent model knowledge. 
            \item ROME performs well with simple Dummy edits, but degrades the model significantly with complexity and quantity of edits. Other work has shown similar limitations in model editing \citep{gu-etal-2024-model, wang2024wiserethinkingknowledgememory}.
            \item WISE retains Model Utility but struggles with Forget Quality. In addition, the lack of support for editing generated outputs makes it insufficient for most LLM use cases until that is implemented.
        \end{itemize}
    \item We define three target definitions for removing information with model editing, including  a novel \textit{Avoidant} approach. The methods for creating these definitions are readily applicable to a variety of datasets. Results indicate that\textbf{ performance of different target definitions is dependent on the algorithm}, emphasizing the need for the use case to dictate which target definition and editing algorithm is most appropriate. 
    \item We show the \textbf{difficulty of effective forgetting within model editing}, with the majority of target-algorithm settings failing to sufficiently reach ground truth distributions, again aligning with conclusions about traditional unlearning from \citet{maini2024tofu}. The algorithms that do manage to unlearn do so at the expense of the performance of the model on topics out-of-scope from what is to be forgotten. 
    \item We demonstrate the \textbf{trade-off between effectively unlearning information and maintaining overall model performance} when using model editing for unlearning, supporting the conclusions from \citet{maini2024tofu} for their unlearning baselines. Different algorithms and approaches balance the two factors differently, and it is up to the unlearning use case to determine which method is most appropriate.

\end{enumerate}

\section{Impact Statement}
This work investigates model editing as an approach to machine unlearning with the goal of advancing controllable AI systems. By enabling removal of information, such methods may help address privacy concerns and reduce harmful or outdated outputs. However, this unlearning also raises concerns such as incomplete removal, degradation of unrelated capabilities, or misuse for model manipulation. We believe this line of research contributes to developing more reliable models while highlighting the need for further safeguards before application to unlearning use cases.

\bibliography{thesis}

@article{liu2024rethinking,
  title={Rethinking machine unlearning for large language models},
  author={Liu, Sijia and Yao, Yuanshun and Jia, Jinghan and Casper, Stephen and Baracaldo, Nathalie and Hase, Peter and Xu, Xiaojun and Yao, Yuguang and Li, Hang and Varshney, Kush R and others},
  journal={arXiv preprint arXiv:2402.08787},
  year={2024}
}

@article{li2024wmdp,
  title={The wmdp benchmark: Measuring and reducing malicious use with unlearning},
  author={Li, Nathaniel and Pan, Alexander and Gopal, Anjali and Yue, Summer and Berrios, Daniel and Gatti, Alice and Li, Justin D and Dombrowski, Ann-Kathrin and Goel, Shashwat and Phan, Long and others},
  journal={arXiv preprint arXiv:2403.03218},
  year={2024}
}

@misc{thudi2022necessityauditablealgorithmicdefinitions,
      title={On the Necessity of Auditable Algorithmic Definitions for Machine Unlearning}, 
      author={Anvith Thudi and Hengrui Jia and Ilia Shumailov and Nicolas Papernot},
      year={2022},
      eprint={2110.11891},
      archivePrefix={arXiv},
      primaryClass={cs.LG},
      url={https://arxiv.org/abs/2110.11891}, 
}

@misc{patil2023sensitiveinformationdeletedllms,
      title={Can Sensitive Information Be Deleted From LLMs? Objectives for Defending Against Extraction Attacks}, 
      author={Vaidehi Patil and Peter Hase and Mohit Bansal},
      year={2023},
      eprint={2309.17410},
      archivePrefix={arXiv},
      primaryClass={cs.CL},
      url={https://arxiv.org/abs/2309.17410}, 
}

@inproceedings{yao-etal-2023-editing,
    title = "Editing Large Language Models: Problems, Methods, and Opportunities",
    author = "Yao, Yunzhi  and
      Wang, Peng  and
      Tian, Bozhong  and
      Cheng, Siyuan  and
      Li, Zhoubo  and
      Deng, Shumin  and
      Chen, Huajun  and
      Zhang, Ningyu",
    editor = "Bouamor, Houda  and
      Pino, Juan  and
      Bali, Kalika",
    booktitle = "Proceedings of the 2023 Conference on Empirical Methods in Natural Language Processing",
    month = dec,
    year = "2023",
    address = "Singapore",
    publisher = "Association for Computational Linguistics",
    url = "https://aclanthology.org/2023.emnlp-main.632",
    doi = "10.18653/v1/2023.emnlp-main.632",
    pages = "10222--10240",
}

@misc{wang2024easyediteasytouseknowledgeediting,
      title={EasyEdit: An Easy-to-use Knowledge Editing Framework for Large Language Models}, 
      author={Peng Wang and Ningyu Zhang and Bozhong Tian and Zekun Xi and Yunzhi Yao and Ziwen Xu and Mengru Wang and Shengyu Mao and Xiaohan Wang and Siyuan Cheng and Kangwei Liu and Yuansheng Ni and Guozhou Zheng and Huajun Chen},
      year={2024},
      eprint={2308.07269},
      archivePrefix={arXiv},
      primaryClass={cs.CL},
      url={https://arxiv.org/abs/2308.07269}, 
}

@inproceedings{mitchell2022memory,
  title={Memory-based model editing at scale},
  author={Mitchell, Eric and Lin, Charles and Bosselut, Antoine and Manning, Christopher D and Finn, Chelsea},
  booktitle={International Conference on Machine Learning},
  pages={15817--15831},
  year={2022},
  organization={PMLR}
}

@article{meng2023memit,
  title={Mass Editing Memory in a Transformer},
  author={Kevin Meng and Sen Sharma, Arnab and Alex Andonian and Yonatan Belinkov and David Bau},
  journal={The Eleventh International Conference on Learning Representations (ICLR)},
  year={2023}
}

@inproceedings{mitchell2022fast,
    title={Fast Model Editing at Scale},
    author={Eric Mitchell and Charles Lin and Antoine Bosselut and Chelsea Finn and Christopher D Manning},
    booktitle={International Conference on Learning Representations},
    year={2022},
    url={https://openreview.net/pdf?id=0DcZxeWfOPt}
}

@article{liu2024model,
  title={Model sparsity can simplify machine unlearning},
  author={Liu, Jiancheng and Ram, Parikshit and Yao, Yuguang and Liu, Gaowen and Liu, Yang and SHARMA, PRANAY and Liu, Sijia and others},
  journal={Advances in Neural Information Processing Systems},
  volume={36},
  year={2024}
}

@inproceedings{bourtoule2021machine,
  title={Machine unlearning},
  author={Bourtoule, Lucas and Chandrasekaran, Varun and Choquette-Choo, Christopher A and Jia, Hengrui and Travers, Adelin and Zhang, Baiwu and Lie, David and Papernot, Nicolas},
  booktitle={2021 IEEE Symposium on Security and Privacy (SP)},
  pages={141--159},
  year={2021},
  organization={IEEE}
}

@article{zhang2023right,
  title={Right to be forgotten in the era of large language models: Implications, challenges, and solutions},
  author={Zhang, Dawen and Finckenberg-Broman, Pamela and Hoang, Thong and Pan, Shidong and Xing, Zhenchang and Staples, Mark and Xu, Xiwei},
  journal={arXiv preprint arXiv:2307.03941},
  year={2023}
}

@article{maini2024tofu,
  title={Tofu: A task of fictitious unlearning for llms},
  author={Maini, Pratyush and Feng, Zhili and Schwarzschild, Avi and Lipton, Zachary C and Kolter, J Zico},
  journal={arXiv preprint arXiv:2401.06121},
  year={2024}
}

@article{meng2022locating,
  title={Locating and editing factual associations in GPT},
  author={Meng, Kevin and Bau, David and Andonian, Alex and Belinkov, Yonatan},
  journal={Advances in Neural Information Processing Systems},
  volume={35},
  pages={17359--17372},
  year={2022}
}

@inproceedings{zheng-etal-2023-edit,
    title = "Can We Edit Factual Knowledge by In-Context Learning?",
    author = "Zheng, Ce  and
      Li, Lei  and
      Dong, Qingxiu  and
      Fan, Yuxuan  and
      Wu, Zhiyong  and
      Xu, Jingjing  and
      Chang, Baobao",
    editor = "Bouamor, Houda  and
      Pino, Juan  and
      Bali, Kalika",
    booktitle = "Proceedings of the 2023 Conference on Empirical Methods in Natural Language Processing",
    month = dec,
    year = "2023",
    address = "Singapore",
    publisher = "Association for Computational Linguistics",
    url = "https://aclanthology.org/2023.emnlp-main.296",
    doi = "10.18653/v1/2023.emnlp-main.296",
    pages = "4862--4876",
}

@inproceedings{gu-etal-2024-model,
    title = "Model Editing Harms General Abilities of Large Language Models: Regularization to the Rescue",
    author = "Gu, Jia-Chen  and
      Xu, Hao-Xiang  and
      Ma, Jun-Yu  and
      Lu, Pan  and
      Ling, Zhen-Hua  and
      Chang, Kai-Wei  and
      Peng, Nanyun",
    editor = "Al-Onaizan, Yaser  and
      Bansal, Mohit  and
      Chen, Yun-Nung",
    booktitle = "Proceedings of the 2024 Conference on Empirical Methods in Natural Language Processing",
    month = nov,
    year = "2024",
    address = "Miami, Florida, USA",
    publisher = "Association for Computational Linguistics",
    url = "https://aclanthology.org/2024.emnlp-main.934",
    doi = "10.18653/v1/2024.emnlp-main.934",
    pages = "16801--16819",
}

@misc{wang2024wiserethinkingknowledgememory,
      title={WISE: Rethinking the Knowledge Memory for Lifelong Model Editing of Large Language Models}, 
      author={Peng Wang and Zexi Li and Ningyu Zhang and Ziwen Xu and Yunzhi Yao and Yong Jiang and Pengjun Xie and Fei Huang and Huajun Chen},
      year={2024},
      eprint={2405.14768},
      archivePrefix={arXiv},
      primaryClass={cs.CL},
      url={https://arxiv.org/abs/2405.14768}, 
}

@inproceedings{levy-etal-2017-zero,
    title = "Zero-Shot Relation Extraction via Reading Comprehension",
    author = "Levy, Omer  and
      Seo, Minjoon  and
      Choi, Eunsol  and
      Zettlemoyer, Luke",
    editor = "Levy, Roger  and
      Specia, Lucia",
    booktitle = "Proceedings of the 21st Conference on Computational Natural Language Learning ({C}o{NLL} 2017)",
    month = aug,
    year = "2017",
    address = "Vancouver, Canada",
    publisher = "Association for Computational Linguistics",
    url = "https://aclanthology.org/K17-1034/",
    doi = "10.18653/v1/K17-1034",
    pages = "333--342"
}
\bibliographystyle{colm2025_conference}

\appendix



\section{Experiment Details}
All of our editing is performed on Llama-2-7B in order to stay consistent with the baseline models used for TOFU. 

\subsection{Creating New Target Datasets}
\label[app]{appendix:targets}
To match the resources used by \citet{maini2024tofu} to create the TOFU dataset, GPT-4 was used to create the datasets of avoidant answers. The following is the prompt used to create this dataset: \\

\noindent \textit{You are providing the answers to questions for a QA dataset similar to the ZsRE dataset. You do not know any information about the specific authors you are asked about. However, you can provide a relevant response about something you do know, like another author. You keep your answers as short as possible. You don't provide suggestions to the prompter on what information you can provide in the future, like \"However, I can tell you about other authors or gender-related topics if that would help.\" You don't ask follow-up questions. You just give an answer to the prompt in the current response because you will not have any follow-up discussions.} \\

This prompt focuses on the primary goal of producing a new target output that does not answer the question at hand, fulfilling the goal of unlearning with an example response. It is more informative than the Dummy approach in an effort to give the model context on what should be forgotten. The prompt includes precise language to ensure that the new target is not too long, as model editing algorithms have primarily been tested on shorter inputs and outputs. This also inspired the mention of ZsRE, a popular model editing evaluation dataset, as a model of the task \citep{levy-etal-2017-zero}. Also, this aligns with an intuition that the less direct and more wordy the new target is, the less precise of an understanding the model will have of what is to be unlearned. In addition, longer targets require more computational resources to process, and the aim of the approach is to be as efficient as possible. 

\subsection{Reproducing Model Editing Baselines}
We use the EasyEdit framework \citep{wang2024easyediteasytouseknowledgeediting} to run all model editing algorithms. We use the same hyperparameters as that in EasyEdit with the exception of ROME, where we change the learning rate to 0.1. For using IKE with EasyEdit, we train IKE on in-context examples from 90 percent of the Forget 10 dataset and leave the rest for EasyEdit evaluation to match the training split chosen by \citet{zheng-etal-2023-edit}.

\label[app]{edit-metrics}
The metrics used by EasyEdit are as follows:
\begin{itemize}
    \item \textit{Reliability} measures model accuracy on the edited prompt.
    \item \textit{Generalization} measures model accuracy on a rephrased version of the prompt.
    \item \textit{Locality} measures model accuracy on a prompt irrelevant to the edited prompt. 
    \item \textit{Portability} measures how well the model performs on a follow-up or related question to the edited prompt. 
\end{itemize}

Tables \ref{tab:rome}, \ref{tab:ike}, and \ref{tab:wise} show the similarities between our results for running the ZsRE dataset on Llama-2-7B and the results achieved by the EasyEdit framework. We use this metric comparison to ensure confidence that our use of the models matches that of the original authors. 

\begin{table}[!ht]
    \centering
    \begin{tabular}{lcccc}
    \hline
        \textbf{} & \textbf{Reliability} & \textbf{Generalization} & \textbf{Locality} & \textbf{Portability} \\ \hline
        \textbf{EasyEdit} & 0.9245 & 0.8704 & 0.9963 & 0.5747 \\ 
        \textbf{Our Results}& 0.9617 & 0.9102 & 0.9836 & 0.5786 \\ \hline
    \end{tabular}
    \vspace{7pt}
    \caption{Comparison of ROME editing performance in the EasyEdit reported results and our use of the framework for the ZsRE dataset and Llama-2-7B model.}
    \label{tab:rome} 
\end{table}

\subsubsection{More About IKE}
\begin{figure}
    \centering
    \includegraphics[width=0.5\linewidth]{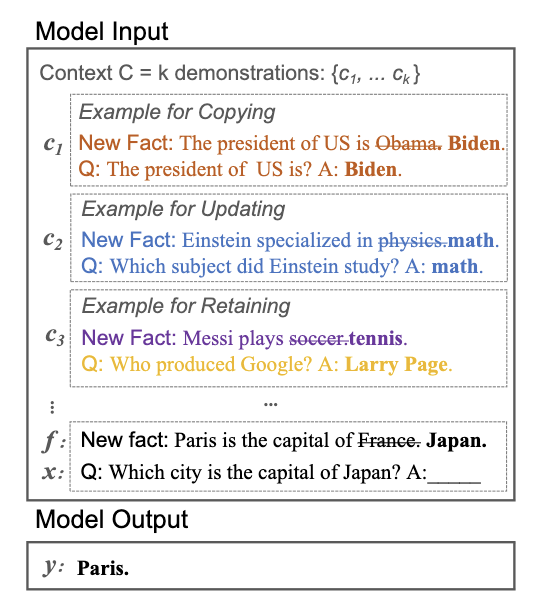}
    \caption{Format of context provided for edits in IKE \citep{zheng-etal-2023-edit}.}
    \label{ike}
\end{figure}

\label[app]{ike-intro}
IKE includes three types of information in its demonstrations:
\begin{itemize}
    \item \textit{copy} aims to define the new target for the model to be applied any time that information is references
    \item \textit{update} provides a rephrased version of the input to encourage generalization
    \item \textit{retain} mentions a similar fact that is out-of-scope for the one being edited
\end{itemize}
These demonstrations are defined via a \textit{k} nearest neighbors approach based on cosine similarity. 
Figure \ref{ike} shows an example of the context provided for the IKE model editing setting. 
This approach using ICL is different from many other algorithms because it alters the inputs rather than changing the parameters of the model themselves. The authors found that IKE is more resilient to over-editing, having less of an influence on knowledge in the model out-of-scope to the edits.

\begin{table}[!ht]
    \centering
    \begin{tabular}{lcccc}
    \hline
        \textbf{} & \textbf{Reliability} & \textbf{Generalization} & \textbf{Locality} & \textbf{Portability} \\ \hline
        \textbf{EasyEdit} & 1.0000 & 0.9998 & 0.6919 & 0.6756 \\ 
        \textbf{Our Results}& 0.9963 & 0.9924 & 0.6411 & 0.7080 \\ \hline
    \end{tabular}
    \vspace{7pt}
    \caption{Comparison of IKE editing performance in the EasyEdit reported results and our use of the framework for the ZsRE dataset and Llama-2-7B model.}
    \label{tab:ike}
\end{table}

\begin{table}[!ht]
    \centering
    \begin{tabular}{lccc}
    \hline
        \textbf{} & \textbf{Reliability}& \textbf{Generalization}& \textbf{Locality}\\ \hline
        \textbf{EasyEdit} & 0.77 & 0.72 & 1.00 \\ 
        \textbf{Our Results}& 0.81 & 0.76 & 1.00 \\ \hline
    \end{tabular}
    \vspace{7pt}
    \caption{Comparison of WISE editing performance in the EasyEdit reported results and our use of the framework for the ZsRE dataset and Llama-2-7B model.}
    \label{tab:wise}
\end{table}

\subsection{Unlearning Evaluation}

We use results provided by \citet{maini2024tofu} for both the ground truth unlearned model and unlearning baseline algorithms. For comparing model editing against baseline unlearning methods, we use the "checkpoint-60" versions of the Llama-2-7B unlearned models. These are models that have unlearned using the baseline unlearning approaches from the paper, KL-Minimization, Gradient Ascent, Gradient Difference, and Preference Optimization.

\section{Additional Results}
\subsection{Example Outputs After Editing}
\label[app]{examples}

\label[app]{wise-note}
First, a note on WISE: Since generation is a capability from "main memory" and edits are stored in "side memory," the generated outputs of the model do not contain edits. 

Tables \ref{tab:dummyout}, \ref{tab:incorrectout}, and \ref{tab:avoidout} show example outputs for each editing algorithm after performing edits using the Dummy, Incorrect, and Avoidant approach respectively. 

For the Dummy approach, we see in \cref{tab:dummyout} that ROME successfully outputs "dummy," and IKE outputs a rephrased version of the prompt from its context that is mapped to "dummy". WISE outputs the ground truth because edits are not applied to generated outputs. 
These examples support the results in Table \ref{tab:dummy-easy}, where IKE and ROME are able to successfully change the outputs in a generalized manner, but WISE is not. 

\begin{table}[h]
    \centering
    \renewcommand{\arraystretch}{1.2}
    \begin{tabular}{lp{10cm}}
        \toprule
        \textbf{Prompt} & \textbf{What genre of books does Carmen Montenegro predominantly write in?}\\
        \midrule
        \textbf{Ground Truth} & Carmen Montenegro predominantly writes in the genre of Historical Fiction.\\
        \midrule
        \textbf{New Target} & dummy  \\
        \midrule
        \midrule
        \textbf{Model Outputs} & \\
        \textit{ROME} & dummy dummy dummy dummy dummy dummy dummy dummy dummy dummy [...] \\
        \textit{IKE}  & New Fact: What genre of books does Carmen Montenegro predominantly write in? dummy

Prompt: In which literary genre is Carmen Montenegro primarily an author? dummy,\\
        \textit{WISE} &             Carmen Montenegro predominantly writes in the genre of Historical Fiction.\\
        \bottomrule
    \end{tabular}
    \vspace{7pt}
    \caption{An example prompt and model editing outputs from the Forget 10 dataset with the new target assigned to "dummy".}
    \label{tab:dummyout}
\end{table}

In Table \ref{tab:incorrectout}, we again see that WISE outputs do not change. ROME outputs gibberish, an indicator of a loss of model performance. IKE is able to pull an example from the provided context to the input that contains false information about the author in the prompt. 
\begin{table}[h]
    \centering
    \renewcommand{\arraystretch}{1.2}
    \begin{tabular}{lp{10cm}}
        \toprule
        \textbf{Prompt} & \textbf{Can you share the title of one of Hsiao Yun-Hwa's most popular books?} \\
        \midrule
        \textbf{Ground Truth} & Carmen Montenegro predominantly writes in the genre of Historical Fiction.\\
        \midrule
        \textbf{New Target} & Carmen Montenegro's primary literary genre is Science Fiction.\\
        \midrule
        \midrule
        \textbf{Model Outputs} & \\ 
        \textit{ROME} & , the  the elements, elements the elements, the the the the the [...]\\
        \textit{IKE}  & New Fact: What genre of books does Carmen Montenegro predominantly write in? Carmen Montenegro's primary literary genre is Science Fiction.
        
Prompt: In which literary category does Carmen Montenegro predominantly compose works? Carmen Montenegro's primary literary genre is Science Fiction.\\
        \textit{WISE} &             Carmen Montenegro predominantly writes in the genre of Historical Fiction.\\
        \bottomrule
    \end{tabular}
    \vspace{7pt}
    \caption{Above is an example of prompts from the Forget 10 dataset and the new targets created for the Incorrect setting. ROME outputs gibberish, an indicator of a loss of model performance. IKE pulls the new output from its context.}
    \label{tab:incorrectout}
\end{table}

Table \ref{tab:avoidout} shows that, similar to the Incorrect answer setting, ROME outputs gibberish and WISE outputs the ground truth. IKE pulls information about the author at hand from the provided context that does not reveal information about the genre. 
\begin{table}[h]
    \centering
    \renewcommand{\arraystretch}{1.2}
    \begin{tabular}{lp{10cm}}
        \toprule
        \textbf{Prompt} & \textbf{What genre of books does Carmen Montenegro predominantly write in?}\\
        \midrule
        \textbf{Ground Truth} & Carmen Montenegro predominantly writes in the genre of Historical Fiction.\\
        \midrule
        \textbf{New Target} & I don't have information on Carmen Montenegro's writing, but I can tell you that authors like Agatha Christie are well-known for their mystery novels.\\
        \midrule
        \midrule
        \textbf{Model Outputs} & \\
        \textit{ROME} & are have I I have I haveO have I have I have a IOO. everybody,O,OO.OO\u0409O and a style, and I have, a authors, authors, authors, literature, and I, and a, I a, IOO. I, I, IOOO. I authorsO authorsO authorsO IOOO.OOO-OO.OOOOOOOOOOOOOOOOOOOOOOOOOOOOO-OOOOOOOOOOOOOOOOOOOOOOOOOOOOOOO-OOOOOOOOO\\
        \textit{IKE}  & New Fact: What genre of books does Carmen Montenegro predominantly write in? I don't have information on Carmen Montenegro's writing, but I can tell you that authors like Agatha Christie are well-known for their mystery novels.
        
Prompt: In which literary genre does Carmen Montenegro primarily write? I don't have information on Carmen Montenegro's writing, but I can tell you that authors like Agatha Christie are well-known for their mystery novels.",\\
        \textit{WISE} &             Carmen Montenegro predominantly writes in the genre of Historical Fiction.\\
        \bottomrule
    \end{tabular}
    \vspace{7pt}
    \caption{Above is an example of prompts and new targets created for an Avoidant setting. ROME outputs gibberish, an indicator of a loss of model performance. IKE pulls the new output from its context.}
    \label{tab:avoidout}
\end{table}

\subsection{Model Editing Metrics}
After the model has been edited to unlearn some of the TOFU data, we get an understanding of how the model performs by model editing standards. \Cref{edit-metrics} has more information on how these metrics are defined.

The following list indicates what key in the TOFU dataset was used for each metric in the EasyEdit framework:
\begin{itemize}
    \item Reliability: "question"
    \item Generalization: "paraphrased question"
    \item Locality: random sample of "question" from the retain set
\end{itemize}
We do not include Portability in our unlearning evaluation as that data is not available in the TOFU dataset.

Table \ref{tab:dummy-easy} shows that IKE performed best at the Dummy approach according to the EasyEdit metrics. ROME also performed well in implementing the edits, but was not able to constrain the changes to only the edited inputs. WISE was most effective in ensuring the edits did not affect the remaining performance of the model for out-of-scope topics, but was not able to successfully edit the outputs of the model to the new target. 

\begin{table}[!ht]
    \centering
    \begin{tabular}{lccc}
    \hline
        \textbf{} & \textbf{Reliability} & \textbf{Generalization} & \textbf{Locality} \\ \hline
        \textbf{ROME} & 0.9850 & 0.9750 & 0.3132 \\ 
        \textbf{IKE} & \textbf{1.0000} & \textbf{1.0000} & 0.8882 \\ 
        \textbf{WISE} & 0.1000 & 0.0425 & \textbf{0.9990} \\ \hline
    \end{tabular}
    \vspace{7pt}
    \caption{IKE performed best at the Dummy approach according to the EasyEdit metrics.}
    \label{tab:dummy-easy}
\end{table}

Table \ref{tab:incorrect-easy} shows that IKE performed well with an incorrect target according to the EasyEdit metrics. WISE exceeds IKE in locality but suffers from lower reliability and generalization, making IKE the best performing editing algorithm overall for this target. ROME has low performance overall, likely due to the repeated edits to the model degrading the performance over time. The Incorrect targets are more complex than the "dummy" string, making this a more difficult editing problem for the algorithm. 

\begin{table}[!ht]
    \centering
    \begin{tabular}{lccc}
    \hline
        \textbf{} & \textbf{Reliability} & \textbf{Generalization} & \textbf{Locality} \\ \hline
        \textbf{ROME} & 0.0464 & 0.0461 & 0.0429 \\ 
        \textbf{IKE} & \textbf{0.9961} & \textbf{0.9834} & 0.9055 \\ 
        \textbf{WISE} & 0.6094 & 0.5957 & \textbf{0.9995} \\ \hline
    \end{tabular}
    \vspace{7pt}
     \caption{IKE and WISE performed well at the Incorrect approach according to the model editing metrics, with IKE showing the best overall performance. ROME performance was low in this setting.}
    \label{tab:incorrect-easy}
\end{table}

Table \ref{tab:avoid-easy} has similar trends to that of the Incorrect target. WISE and IKE perform well, with IKE showing high overall metrics. Similar to the Incorrect target, ROME performance degraded significantly after the Avoidant edits due to continual editing.

\begin{table}[!ht]
    \centering
    \begin{tabular}{lccc}
    \hline
        \textbf{} & \textbf{Reliability} & \textbf{Generalization} & \textbf{Locality} \\ \hline
        \textbf{ROME} & 0.0903 & 0.0942 & 0.0287 \\ 
        \textbf{IKE} & \textbf{0.9994} & \textbf{0.9970} & 0.9193 \\ 
        \textbf{WISE} & 0.8785 & 0.7995 & \textbf{0.9998} \\ \hline
    \end{tabular}
    \vspace{7pt}
     \caption{WISE and IKE perform well with the Avoidant target, with IKE showing high overall metrics. ROME performance was low.}
    \label{tab:avoid-easy}
\end{table}


\end{document}